\documentclass[letterpaper]{article} % DO NOT CHANGE THIS
\usepackage{aaai25}  % DO NOT CHANGE THIS
\usepackage{times}  % DO NOT CHANGE THIS
\usepackage{helvet}  % DO NOT CHANGE THIS
\usepackage{courier}  % DO NOT CHANGE THIS
\usepackage[hyphens]{url}  % DO NOT CHANGE THIS
\usepackage{graphicx} % DO NOT CHANGE THIS
\usepackage{natbib}  % DO NOT CHANGE THIS AND DO NOT ADD ANY OPTIONS TO IT 
\usepackage{caption} % DO NOT CHANGE THIS AND DO NOT ADD ANY OPTIONS TO IT
\usepackage{booktabs}
\usepackage{amsmath}
\usepackage{dsfont}
\usepackage{algorithm}
\usepackage{algorithmic}

\usepackage{hhline}
\usepackage{multirow}
\usepackage{makecell}
\usepackage{paralist}

\usepackage{tablefootnote}
\newcommand{\footnoteagain}[1]{\footnotemark[\getrefnumber{#1}]}
\usepackage{newfloat}
\usepackage{listings}
\DeclareCaptionStyle{ruled}{labelfont=normalfont,labelsep=colon,strut=off} % DO NOT CHANGE THIS
\floatstyle{ruled}
\newfloat{listing}{tb}{lst}{}
\floatname{listing}{Listing}
\usepackage{tikz-dependency}
\usepackage{subcaption}
\usepackage{refcount}

\title{Error Diversity Matters: An Error-Resistant Ensemble Method\\for Unsupervised Dependency Parsing}
\author {
    Behzad Shayegh\textsuperscript{\rm 1},
    Hobie H.-B. Lee\textsuperscript{\rm 1},
    Xiaodan Zhu\textsuperscript{\rm 2},
    Jackie Chi Kit Cheung\textsuperscript{\rm 3,4},
    Lili Mou\textsuperscript{\rm 1,4}
}
\affiliations {
    \textsuperscript{\rm 1}Dept. Computing Science, Alberta Machine Intelligence Institute (Amii), University of Alberta\\
    \textsuperscript{\rm 2}Dept. Electrical and Computer Engineering \& Ingenuity Labs Research Institute, Queen’s University\\
    \textsuperscript{\rm 3}Quebec Artificial Intelligence Institute (Mila), McGill University\\
    \textsuperscript{\rm 4}Canada CIFAR AI Chair\\
    the.shayegh@gmail.com, 
    hobie.lee@ualberta.ca, \\
    xiaodan.zhu@queensu.ca, 
    jcheung@cs.mcgill.ca, 
    doublepower.mou@gmail.com
}

\begin{document}

\maketitle

\begin{abstract}
    We address unsupervised dependency parsing by building an ensemble of diverse existing models through post hoc aggregation of their output dependency parse structures. We observe that these ensembles often suffer from low robustness against weak ensemble components due to error accumulation. To tackle this problem, we propose an efficient ensemble-selection approach that considers error diversity and avoids error accumulation. Results demonstrate that our approach outperforms each individual model as well as previous ensemble techniques. Additionally, our experiments show that the proposed ensemble-selection method significantly enhances the performance and robustness of our ensemble, surpassing previously proposed strategies, which have not accounted for error diversity.
\end{abstract}

\begin{links}
    \link{Code}{https://github.com/MANGA-UOFA/ED4UDP}
    % \link{Published version}{https://aai/???}
\end{links}

\section{Introduction}
\label{sec:intro}

Syntactic parsing, a fundamental task in natural language processing (NLP), refers to identifying grammatical structures in text~\cite{zhang2020survey}, which can help downstream NLP tasks such as developing explainable models~\citep{amara2024syntaxshap}. Unsupervised syntactic parsing is particularly beneficial for processing languages or domains with limited resources, as it eliminates the need for human-annotated training data~\citep{kann-etal-2019-neural}. In addition, autonomously discovering patterns and structures helps to provide evidence to test linguistic theories~\citep{goldsmith2001unsupervised} and cognitive models~\citep{Exemplar2Grammar}. Two common types of syntactic parsing include: (1) constituency parsing, organizing words and phrases in a sentence into a hierarchy~\citep{constituencyTree}, and (2) dependency parsing, predicting dependence links between words in a sentence~\citep{tesniere1959elements}. The latter is particularly interesting given its possible applications to other fields, e.g., RNA structure prediction~\citep{anonymous2024depfold}.

Unsupervised dependency parsing has seen diverse approaches over decades~\citep{smith2005guiding,cai2017crf,han-etal-2020-survey}. \citet{yang-etal-2020-second} show that the combination of two different models can outperform both individuals, suggesting the varied contributions of different models.
We follow our previous work on constituency parsing~\citep{shayegh2023ensemble,shayegh2024disco} and build an ensemble of different unsupervised dependency parsers to leverage their diverse expertise.\footnote{Expertise diversity may be due to different inductive biases, objective functions, data features~\citep{ensemblebook}, and/or different training subsets~\citep{breiman1996bagging,wen2023equalsizehardemalgorithm,wen2024ebbs}. In our work, we build an ensemble of different models exploring various structures, predictors, and training methods.} We regard the ensemble output as the dependency structure that maximizes its overall similarity to all the \textit{individual} models' outputs. For the similarity metric, we employ the unlabeled attachment score~\citep[UAS;][]{nivre2017universal}, which is the most widely used evaluation metric of the task.

\begin{figure}[t]
    \centering
    \includegraphics[width=\linewidth]{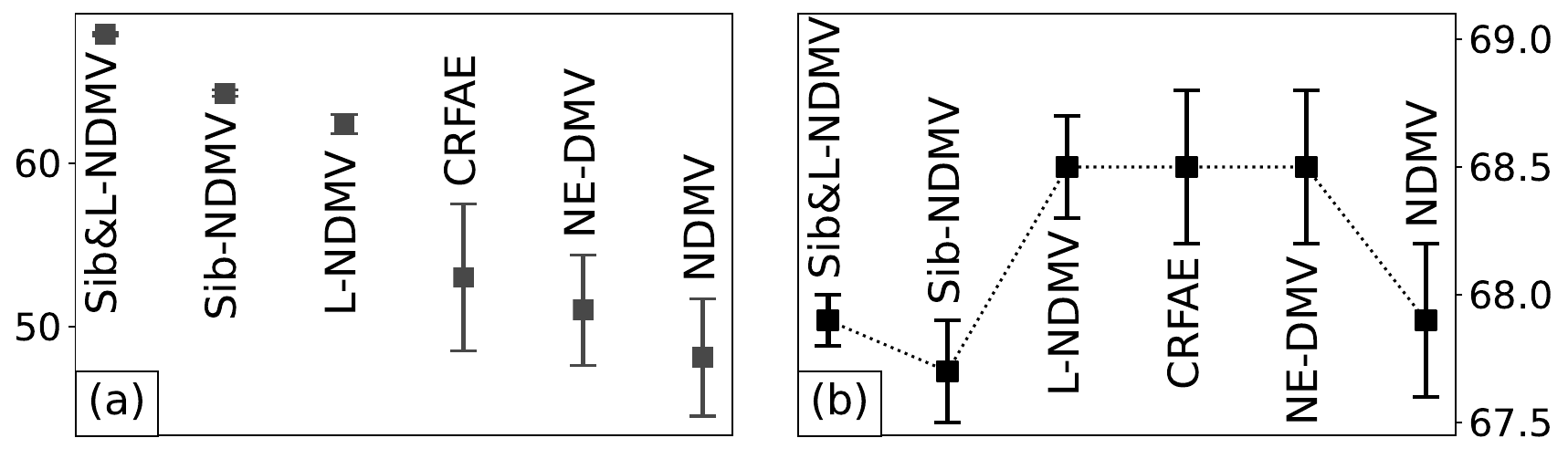}
    \caption{(a) Performances of individuals. (b) Performances of ensembles by adding individuals in (a) from left to right. Numbers are UASs on the WSJ test set, reported by the mean and the standard deviation through five runs. The experiment setup is detailed in \S\ref{sec:settings}.}
    \label{fig:best_to_worst}
\end{figure}

In our work, we observe that a na\"{\i}ve application of \citet{shayegh2023ensemble}'s ensemble is sensitive to weak individuals in dependency parsing. As shown in Figure \ref{fig:best_to_worst}, a best-to-worst incremental ensemble experiment encounters a significant drop in performance when weaker individuals are added. This drop is due to the accumulation of individual errors, which arises from a low error diversity.

It is important to distinguish between two types of diversity: \textit{expertise diversity} and \textit{error diversity}. The former refers to the phenomenon that different models excel on different subsets of samples, while the latter indicates that models are wrong in different ways for each sample. Although both types of diversity are crucial for successful ensembles~\citep{zang2012ensemble}, it is essential to emphasize that expertise diversity serves as a motivation to build an ensemble, whereas error diversity is a requirement for its success (outperforming ensemble individuals), because otherwise, the individuals' errors cannot be eliminated through the ensemble process.

To this end, we propose a diversity-aware method for ensemble selection considering error diversity so that our approach is resistant to error accumulation. Having a diverse set of ensemble components has been a focus of researchers for years~\citep{tang2006analysis,zhou2012ensemble,wu2021boosting, purucker2023qdoes}. However, most previous studies focus on expertise diversity by considering only the individuals' successes/failures, ignoring the differences in their mistakes~\citep{yule1900vii,fleiss1981statistical,cunningham2000diversity}. By contrast, we develop a metric, called \textit{society entropy}, to capture both aspects of diversity. Our approach outperforms the existing expertise-diversity metrics in ensemble selection, demonstrating the importance of error diversity.

We conduct our experiments on the Wall Street Journal (WSJ) corpus in the Penn Treebank~\citep{marcus1993building}. Results show the effectiveness of our ensemble approach, outperforming all the individual parsers and ensemble baselines. We further boost the performance by employing our diversity-aware selection of individuals, achieving state-of-the-art performance in the task.

In conclusion, our main contributions include:
\begin{enumerate}
    \item proposing an ensemble-based approach to unsupervised dependency parsing,
    \item specifying error diversity as an important aspect in building an ensemble, and 
    \item utilizing society entropy as a diversity metric to perform selection of ensemble individuals.
\end{enumerate}

\section{Related Work}

\subsection{Unsupervised Dependency parsing}
Over the years, researchers have proposed various models to tackle unsupervised dependency parsing under different setups~\citep{le2015unsupervised,he2018unsupervised,shen2020structformer}. The longest line of research began with \citet{klein-manning-2004-corpus} introducing the dependency model with valence (DMV), a probabilistic generative model of a sentence and its dependency structure, which is extended by \citet{headden2009improving} to include categories of valence and lexical information. In the deep learning era, \citet{jiang2016unsupervised} utilize neural networks to learn probability distributions in DMV. \citet{han-etal-2017-dependency,han-etal-2019-enhancing,HAN2019105} further equip neural DMVs with lexical and contextual information. More recently, \citet{yang-etal-2020-second} introduce two second-order variants of DMV, which incorporate grandparent--child or sibling information, respectively. In this work, we build ensembles of different models to leverage their diverse knowledge.

All of the above work focuses on projective dependency parse structures, meaning that all the dependency arcs can be drawn on one side of the sentence without crossing each other. The projectivity constraint comes from the context-free nature of human languages~\citep{chomsky1956three,gaifman1965dependency}. Given that most English sentences are characterized by projective structures~\citep{wu2020perturbed}, we also follow previous studies and focus on this type of parses.

Unsupervised dependency parsing typically considers unlabeled structures, i.e., dependency arcs are not categorized by their linguistic types (e.g., subject or object of a verb). We also follow the previous line of research and focus on unlabeled structures.

\subsection{Ensemble-Based Dependency Parsing}
\citet{che2018towards} and \citet{lim-etal-2018-sex} propose to build ensembles of dependency parsers to smooth out their noise and reduce the sensitivity to the initialization of the neural networks. They use the average of networks' softmax outputs for prediction. This approach is restricted as it requires the models to have the same output space, which does not hold in our scenario as we aim to leverage the knowledge of diverse models with different architectures.

\citet{kulkarniCPTAM} and \citet{shayegh2023ensemble} show the effectiveness of post hoc aggregation methods for ensemble-based supervised and unsupervised constituency parsing. \citet{kulkarni2024empirical} borrow different post hoc aggregation methods from graph studies, including the maximum spanning tree~\citep{gavril1987generating}, conflict resolution on heterogeneous data~\citep{li2014resolving}, and a customized Ising model~\citep[CIM;][]{ravikumar2010high}, and employ them for ensemble-based dependency parsing. They compare the performance of these aggregation methods and show the superiority of CIM. However, this method does not ensure the validity of the ensemble output as a projective dependency parse structure. Nevertheless, we include CIM as a baseline which underperforms when applied to our unsupervised setting, performing worse than the best individual.

In this work, we develop a dependency-structure aggregation method, based on which we further build an ensemble of unsupervised dependency parsers with different designs.

\subsection{Ensemble Selection}
\label{sec:ensembleselectionrelatedwork}

Ensemble selection refers to selecting a set of models to form an ensemble, which has been extensively discussed in the machine learning literature~\citep{kuncheva2004combining,caruana2006getting}. \citet{caruana2004ensemble} introduce a forward-stepwise-selection method, inspired by the field of feature selection~\citep{liu2007computational}; in their approach, individuals are incrementally added to the ensemble based on the performance boost on a validation set. Such an approach is prone to overfitting to the validation set~\citep{KOHAVI1997273}; in addition, it is a time-consuming process to build the ensemble, as we need to build many ensembles at each increment.

Another line of research predicts the success of an ensemble by looking at the individuals' properties~\citep{ganaie2022ensemble,mienye2022survey}. One commonly adopted criterion is to keep diversity among the ensemble individuals~\citep{minku2009impact,wood2023unified}. This brings multiple benefits, such as bringing different expertise~\citep[expertise diversity;][]{zang2012ensemble} and smoothing out individual errors~\citep[error diversity;][]{zhou2012ensemble}. Two branches of related studies are: (1) how to make a balance between the diversity and the quality of the individuals~\citep{Chandra2006tradeoff, wood2023unified}, and (2) how to measure the diversity~\citep{kuncheva2003measures}. In our work, we address the latter.

Most ensemble-selection studies focus on binary classification tasks. \citet{yule1900vii} propose a $Q$ statistic to measure diversity based on the association between random variables. \citet{fleiss1981statistical} introduces the measure of inter-rater reliability, quantifying the non-coincidental agreements among classifiers. \citet{kohavi1996bias} measure the diversity based on the variability of predictions across all classifiers. \citet{skalak1996sources} and \citet{ho1998random} measure the disagreement between a pair of classifiers by the ratio of observations on which only one classifier is correct. To address multi-class classification, \citet{kuncheva2003measures} and \citet{kadkhodaei2016entropy} reduce the problem to binary classification by considering the success/failure of classifiers, which is a binary variable. Although this approach measures expertise diversity, it loses information about error diversity, which we show is important and should not be ignored.

\section{Approach}

In this work, we propose an ensemble approach to unsupervised dependency parsing. In general, ensemble methods in machine learning consist of two stages: obtaining individual models and aggregating the individuals' predictions. We will address both in the rest of this section.

\subsection{Aggregating Dependency Parses}
\label{sec:aggs}

We aim to aggregate parses of different individual models, potentially with different architectures and output formats. We propose to have post hoc aggregation methods applied to the dependency parses obtained by previous parsers. Inspired by our previous work \citep{shayegh2023ensemble}, we formulate our ensemble under the minimum Bayes risk (MBR) framework. Consider a set of individuals' outputs $A_1,\cdots, A_K$ given a sentence. The ensemble output $A^*$ is defined as the dependency parse structure maximizing its similarity to all the individuals' outputs:
\begin{align}
    A^* = \underset{A\in \mathcal{A}}{\operatorname{argmax}} \sum_{k=1}^K \operatorname{similarity}(A, A_k)
    \label{eq:MBR}
\end{align}
where $\mathcal{A}$ is the set of all the possible dependency parse structures and $\operatorname{similarity}$ is a customizable similarity measure for dependency parses.

In particular, we propose to use the sentence-level unlabeled attachment score~\citep[UAS;][]{nivre2017universal} as the similarity measure because it is the main evaluation metric for unlabeled, projective dependency parsing.\footnote{As an alternative to UAS, we also follow our previous work~\citep{shayegh2023ensemble} using F\textsubscript{1} as the similarity measure. We show results in Appendix~A. We exclude it from the main body as it underperforms individuals and is considered a failure, demonstrating the importance of using the task-specific evaluation metric as the ensemble objective.}
Formally, UAS is the accuracy of head detection for every word~\citep{le2015unsupervised,nivre2017universal,yang-etal-2020-second}. In other words, $\operatorname{UAS}(A_p, A_r) =~\frac{1}{n} \sum_{j=1}^{n} \mathds{1}[A_p^{(j)}{=}A_r^{(j)}]$, where $A_p$ and $A_r$ are two parallel lists of attachment heads for $n$ given words in a sentence or a corpus. Putting UAS in Eqn.~\eqref{eq:MBR}, we have
\begin{align}
    A^* &= \underset{A\in \mathcal{A}}{\operatorname{argmax}} \sum_{k=1}^K \operatorname{UAS}(A, A_k)\\
    &= \underset{A\in \mathcal{A}}{\operatorname{argmax}} \sum_{j=1}^{n}  \underbrace{\sum_{k=1}^K \mathds{1}\big[A^{(j)}{=}A_k^{(j)}\big]}_{\displaystyle{{m}(A^{(j)};j)}}
\end{align}
Here, ${m}(a; j)$ is the number of individuals assigning word $a$ as the $j$th word's head. Specially, $a=0$ indicates the \textsc{root}.
We solve this problem by reducing it to the dependency parsing model of \citet{eisner-1996-three} and using their efficient dynamic programming algorithm.
\begin{align}
    A^* &= \underset{A\in \mathcal{A}}{\operatorname{argmax}} \sum_{j=1}^{n}  {m}(A^{(j)};j)\\
    &= \underset{A\in \mathcal{A}}{\operatorname{argmax}} \;\exp\Big\{\sum_{j=1}^{n}  {m}(A^{(j)};j)\Big\}\\
    &= \underset{A\in \mathcal{A}}{\operatorname{argmax}} \prod_{j=1}^{n}  \underbrace{\exp\{{m}(A^{(j)};j)\}}_{\displaystyle{\hat{m}(A^{(j)};j)}}\\
    &= \underset{A\in \mathcal{A}}{\operatorname{argmax}} \prod_{j=1}^{n}  \underbrace{\frac{\hat{m}(A^{(j)};j)}{\sum_{a=0}^n \hat{m}(a; j)}}_{\displaystyle{{p}(A^{(j)}; j)}} \label{eq:converttoprob}
\end{align}
In Eqn.~\eqref{eq:converttoprob}, we normalize values of~$\hat{m}$ to be in $[0, 1]$, which can be viewed as estimated probabilities. Consequently, the argmax objective is the joint probability of all attachments in a sentence, i.e., the probability of the dependency parse~$A$. Hereby, we have reduced our problem to the dependency parsing problem in \citet{eisner-1996-three} and can use their efficient algorithm to obtain $A^*$ in $\mathcal{O}(n^3)$ time complexity.

\subsection{Ensemble Selection}
\label{sec:ensembleselectionapproach}

In this work, we build an ensemble of various unsupervised dependency parsers. We further propose to perform ensemble selection, as we notice in Figure~\ref{fig:best_to_worst} that an ensemble including all individuals underperforms a moderately sized ensemble.

We hypothesize that, if ensemble individuals lack diversity in their errors, they introduce systematic bias and may hurt the ensemble performance. Therefore, it is important to balance the quality of individuals and their diversity. Our ensemble selection objective is to find $\mathcal K\subseteq\{1,\cdots, K\}$ being the set of selected ensemble individuals maximizing
\begin{multline}
    \operatorname{objective}({\mathcal K}) = \sum_{\kappa\in\mathcal K} \operatorname{UAS}(A_{\kappa}, A_{\text{gold}}) + \\\alpha \cdot \operatorname{diversity}(\{A_{\kappa}\}_{\kappa\in\mathcal K})
    \label{eq:modelselectionobjective}
\end{multline}
where $A_{\text{gold}}$ is the ground-truth parse\footnote{\label{footnote:labeled_validation}We assume a small labeled validation set is available, which is a common assumption in both unsupervised syntactic structure detection~\citep{jiang2016unsupervised,deshmukh-etal-2021-unsupervised-chunking,10.1162/coli_a_00545} and ensemble selection~\citep{mienye2022survey}.} and $\alpha$ is a balancing hyperparameter.

For $\operatorname{diversity}$, we introduce \textit{society entropy}. Essentially, we first define \textit{society distribution} ($\operatorname{SD}$) for the $j$th word as
\begin{align}
    \operatorname{SD}\Big(a; \{A^{(j)}_{\kappa}\}_{\kappa\in\mathcal K}\Big) = \frac{\sum_{\kappa\in\mathcal K} \mathds{1}\big[A_\kappa^{(j)}{=}a\big]}{|{\mathcal{K}}|}
\end{align}
where the probability of the head being $a$ is the fraction of the individuals agreeing on that. Then, we define the \textit{society entropy} ($\operatorname{SE}$) as the entropy $E(p)=-\sum_x p(x)\log p(x)$ of society distribution:
\begin{align}
    \operatorname{SE}(\{A_{\kappa}\}_{\kappa\in\mathcal K}) = \frac{1}{n} \sum_{j=1}^n E \Big(\operatorname{SD}\Big(\cdot\:; \{A^{(j)}_{\kappa}\}_{\kappa\in\mathcal K}\Big)\Big)\label{eq:socentropy}
\end{align}
where $n$ is the number of words. This metric not only considers expertise diversity, but also measures the error diversity across individuals.

We maximize our objective in a forward-stepwise manner~\cite{caruana2004ensemble}. In other words, we begin by selecting
\begin{equation}
    \kappa_1=\underset{\kappa' \in \{1, \cdots, K\}}{\operatorname{argmax}} \operatorname{UAS}(A_{\kappa'}, A_{\text{gold}})
\end{equation}
Then, we add individuals to our collection one at a time by maximizing our objective, i.e., the $t$th selected individual is chosen by
\begin{equation}
    \kappa_t = \underset{\kappa' \in \{1, \cdots, K\}\backslash\mathcal{K}_{t-1}}{\operatorname{argmax}}\operatorname{objective}(\mathcal{K}_{t-1}\cup\{\kappa'\}) 
\end{equation}
where $\mathcal{K}_{t} = \{\kappa_1,\cdots,\kappa_{t}\}$ is the first $t$ selected individuals. Overall, the selected set of individuals is $\mathcal{K}_T$, where $T$ is a hyperparameter indicating the number of ensemble individuals.

\section{Experiments}

\subsection{Settings}
\label{sec:settings}

\paragraph{Dataset.} We performed experiments on the Wall Street Journal (WSJ) corpus in the Penn Treebank~\citep{marcus1993building}. We adopted the standard split: Sections~2--21 for training, Section~22 for validation, and Section~23 for testing. For training and validation, we followed previous work and only used sentences with at most 10 words after being stripped of punctuation and terminals. We used the entire test set for evaluation, regardless of sentence lengths. As all our individuals are unsupervised parsers, we did not use linguistic annotations in the training set. We use an annotated validation set\footnotemark[3] with only 250 sentences for validation during training and ensemble selection.

\paragraph{Ensemble Individuals.} We consider the following unsupervised dependency parsers as our ensemble components\footnote{Our choice of individuals was highly based on the replicability of the baselines.}:

\begin{itemize}
    \item {CRFAE}~\citep{cai2017crf}, which is a discriminative and globally normalized conditional random field autoencoder model that predicts the conditional distribution of the structure given a sentence.

    \item {NDMV}~\citep{jiang2016unsupervised}, a dependency model with valence~\citep[DMV;][]{klein-manning-2004-corpus} model that learns probability distributions using neural networks. We used Viterbi expectation maximization to train the model.

    \item {NE-DMV}~\citep{jiang2016unsupervised}, which applies the neural approach to the extended DMV model \citep{headden2009improving,gillenwater2010sparsity}. We employed \citet{tu2012unambiguity} for initialization.

    \item {L-NDMV}~\citep{han-etal-2017-dependency}, an extension of NDMV that utilizes lexical information of tokens.

    \item {Sib-NDMV}~\citep{yang-etal-2020-second}, which applies the neural approach to Sib-DMV, an extension of DMV that utilizes information of sibling tokens.

    \item {Sib\&L-NDMV}~\citep{yang-etal-2020-second}, a joint model incorporating Sib-NDMV and L-NDMV.
\end{itemize}
For L-NDMV, Sib-NDMV, and Sib\&L-NDMV, we use \citet{naseem2010using} for initialization.
For the hyperparameters and configurations of individuals, we use the default values specified in the respective papers or codebases. We train five instances of each model using different random seeds to assess the stability of our approach. We observe that CRFAE, NDMV, and NE-DMV exhibit instability and lower performance on average than what the authors reported. To achieve comparable performance, we run the training for these models 20 times and select the top 25\% based on their performance on the validation set. We report a comparison between the quoted results and our replication results in Appendix~B, showing the success of our replication.

\subsection{Aggregation Results}
\label{sec:results}

\begin{table}[t]
    \centering
    \begin{tabular}{|r l | l |}
        \toprule
        & Method                             & UAS                               \\
        \midrule
        \multicolumn{2}{|l|}{Ensemble individuals} & \\
        1  & CRFAE                              & $53.0_{\pm4.5}$                   \\
        2  & NDMV                               & $48.1_{\pm3.6}$                   \\
        3  & NE-DMV                             & $51.0_{\pm3.4}$                   \\
        4  & L-NDMV                             & $62.4_{\pm0.6}$                   \\
        5  & Sib-NDMV                           & $64.3_{\pm0.2}$                   \\
        \midrule
        \multicolumn{2}{|l|}{Ensembles} & \\
        6  & CIM aggregation                    & $58.8_{\pm0.7}$                   \\
        7  & Our unweighted aggregation         & ${{65.7}}_{\pm0.9}^{**}$          \\
        8  & Our weighted aggregation           & ${{66.6}}_{\pm0.4}^{***}$         \\
        \midrule
        \multicolumn{2}{|l|}{Selected ensembles (our aggregation)} & \\
        9  & w/o diversity                      & $63.8$                            \\
        10 & w/ Kuncheva's diversity            & $65.8$                            \\
        11 & w/ society entropy (ours)          & $\underline{67.3}$                \\
        12 & Ensemble validation                & $\bf{67.8}$                       \\
        % \hline
        \midrule
        \midrule
        % \hline
        \multicolumn{2}{|l|}{Additional individual} & \\
        13 & Sib\&L-NDMV                        & $67.9_{\pm0.1}$                   \\
        \midrule
        \multicolumn{2}{|l|}{Ensembles (our aggregation)} & \\
        14 & Unweighted                         & ${{67.9}}_{\pm0.3}$               \\
        15 & Weighted                           & ${\underline{68.4}}_{\pm0.4}^{*}$ \\
        16 & Selected w/ society entropy (ours) & $\bf{68.8}$                       \\
        17 & Selected by ensemble validation    & $\underline{68.4}$                \\
        \bottomrule
    \end{tabular}
    \caption{Mean and standard deviation across five runs, evaluated on the WSJ test set. $^*$, $^{**}$, and $^{***}$ denote statistically significant improvements over the best individual, as determined by a paired-sample T-test at confidence levels of 92\%, 98\%, and 99.99\%, respectively. Selected ensembles do not include variance statistics, and thus, statistical tests can not be conducted for them.}
    \label{tab:main_results}
\end{table}

\begin{table*}[t]
    \centering
    \begin{tabular}{| l || r || c  c  c | c |}
        \toprule
        \multirow{2}{*}{Selection method} & \multirow{2}{*}{Selection time (ms)} & \multicolumn{4}{ c |}{Ensemble UAS} \\
        &                   & $k=3$             & $k=4$               & $k=5$              & Overall              \\
        \midrule
        w/o diversity           & $0.03_{\pm0}$     & $64.0_{\pm0.9}$   & $65.6_{{\pm1.3}}$   & $65.9_{\pm0.7}$    & $65.2_{\pm1.3}$      \\
        w/ Kuncheva's diversity & $12_{\pm004}$     & $65.0_{\pm0.4}$   & ${67.7_{\pm0.5}}$   & $ {66.1}_{\pm0.6}$ & $66.3_{\pm1.2}$\\
        w/ society entropy (ours)     & $22_{\pm006}$     & ${67.1_{\pm0.3}}$ & ${{66.9}_{\pm0.3}}$ & $67.2_{{\pm0.3}}$ & $\bf{67.1_{\pm0.4}}$\\
        Ensemble validation     & $11,066_{\pm166}$ & $66.6_{\pm0.8}$   & $66.6_{\pm0.4}$     & $67.7_{\pm0.5}$ & $67.0_{\pm0.8}$\\
        \bottomrule
    \end{tabular}
    \caption{UASs of selected ensembles with $k$ ensemble components, selected from 15 individuals using different strategies. Numbers are the means and standard deviations across 30 runs. The Overall column represents all the 90 results for $k=3,4,5$. Time is measured using 28 Intel(R) Core(TM) i9-9940X (@3.30GHz) CPUs.}
    \label{tab:model_selection_main}
\end{table*}

Table~\ref{tab:main_results} shows the main results on the WSJ dataset, where we performed five runs of each individual and built ensembles of corresponding runs. We may or may not exclude Sib\&L-NDMV as it is already a fusion model, and one may argue about including it in an ensemble where Sib-NDMV and L-NDMV are also present.

We compare our aggregation approach against the customized Ising model~\citep[CIM; Row~6;][]{ravikumar2010high}, which calculates per-sample vote weights based on individuals' correlation over samples in classification tasks. Indeed, \citet{kulkarni2024empirical}, as the only previous work on post hoc ensemble-based dependency parsing, shows that CIM outperforms alternatives, including the maximum spanning tree~\citep{gavril1987generating} and conflict resolution on heterogeneous data~\citep{li2014resolving}.\footnote{We strictly follow \citet{kulkarni2024empirical} to employ CIM in dependency parsing, using their code from \url{https://github.com/kulkarniadithya/Dependency_Parser_Aggregation}} The CIM is considered an unweighted approach, as it does not use labeled validation data. We see that this approach does not retain the performance of the best individual (Row~5) and is thus unsuccessful.

By contrast, our approach (Row~7) outperforms all the individuals and baselines, demonstrating the effectiveness of our MBR formulation. We also observe high stability in the performance of our ensemble despite the unstable individuals (Rows~1-3). However, we observe that building an ensemble including Sib\&L-NDMV (Row~14) does not bring a performance
boost over the best individual (Row~13) due to the huge performance gap between individuals, showing the limited robustness of the ensemble against weak individuals.

We further improve the performance of our ensemble by weighting individuals based on their performance on the validation set (Rows~8 and~15).
We can build weighted ensembles with rational numbers as weights by virtually duplicating individuals~\citep{shayegh2024disco}.
The weighted variant overcomes the challenge of high variance in individuals' performance, exhibiting higher performance than the best individual and the unweighted ensemble. These results highlight the importance of controlling the negative effect of weak individuals. In the next section, we explore ensemble selection as a superior alternative technique, further enhancing robustness against weak individuals.

\subsection{Ensemble Selection Results}

We compare our diversity-based selection method with the classic forward-stepwise-selection approach~\citep{caruana2004ensemble}, referred to as \textit{ensemble validation}. This baseline validates each possible ensemble on a validation set, and thus is slow and computationally expensive. Moreover, we compare our proposed diversity metric ``society entropy'' versus several previously proposed measures, listed bellow:

\begin{itemize}
    \item {Disagreement}~\citep{skalak1996sources,ho1998random}, which is a pairwise metric, indicating the fraction of times only one classifier is true. We extend this metric to more than two classifiers by averaging disagreements through all pairs:
    \begin{align}
        \sum_{\kappa_1,\kappa_2\in\mathcal{K}} \sum_{j=1}^n \frac{\mathds{1}\big[A_{\kappa_1}^{(j)}{=}A_{\text{gold}}^{(j)}\big] \oplus \mathds{1}\big[A_{\kappa_2}^{(j)}{=}A_{\text{gold}}^{(j)}\big]}{n|\mathcal{K}|^2}
    \end{align}
    where $n$ is the number of samples based on which we compute the diversity, $A_{\text{gold}}$ is the ground truth, $\mathcal{K}$ is the selected individuals, and $\oplus$ is the ``exclusive or'' operator.

    \item {KW variance}~\citep{kohavi1996bias}, which differs from the averaged disagreement metric above by the constant coefficient $\frac{K-1}{2K}$~\citep{kuncheva2003measures}:
    \begin{align}
        \sum_{j=1}^{n}\frac{\Big(\sum\limits_{\kappa\in\mathcal{K}} \mathds{1}\big[A_{\kappa}^{(j)}{=}A_{\text{gold}}^{(j)}\big]\Big)\Big(\sum\limits_{\kappa\in\mathcal{K}} \mathds{1}\big[A_{\kappa}^{(j)}{\neq}A_{\text{gold}}^{(j)}\big]\Big)}{n|\mathcal{K}|^2}
    \end{align}

    \item {Fleiss' Kappa}~\citep{fleiss1981statistical}, which differs from KW variance by a coefficient:
    \begin{align}
        \sum_{j=1}^{n}\frac{\Big(\sum\limits_{\kappa\in\mathcal{K}} \mathds{1}\big[A_{\kappa}^{(j)}{=}A_{\text{gold}}^{(j)}\big]\Big)\Big(\sum\limits_{\kappa\in\mathcal{K}} \mathds{1}\big[A_{\kappa}^{(j)}{\neq}A_{\text{gold}}^{(j)}\big]\Big)}{n|\mathcal{K}|(1-|\mathcal{K}|)\Bar{p}(1-\Bar{p})}
    \end{align}
    where $\Bar{p} = \frac{1}{\mathit{n|\mathcal{K}|}}\sum_{j=1}^n\sum_{\kappa\in\mathcal{K}}\mathds{1}\big[A_{\kappa}^{(j)}{=}A_{\text{gold}}^{(j)}\big]$ is the fraction of true hits by all the individuals.

    \item Kuncheva's diversity\footnote{\citet{kuncheva2003measures} name this measure as \textit{entropy}. We avoid following this naming due to the different formulation of the standard entropy which we use for our society entropy.}~\citep{kuncheva2003measures}, which is a non-pairwise metric measuring the smoothness of the oracle distribution:
    \begin{multline}
        \frac{1}{n (|\mathcal{K}|-\lceil |\mathcal{K}|/2 \rceil)}\sum_{j=1}^{n} \min\bigg\{\sum_{\kappa\in\mathcal{K}}\mathds{1}\big[A_{\kappa}^{(j)}{=}A_{\text{gold}}^{(j)}\big],\\ \sum_{\kappa\in\mathcal{K}} \mathds{1}\big[A_{\kappa}^{(j)}{\neq}A_{\text{gold}}^{(j)}\big]\bigg\}\label{eqn:Kuncheva_diversity}
    \end{multline}

    \item {PCDM}~\citep{banfield2005ensemble}, which is based on the proportion of classifiers getting each sample correct:
    \begin{align}
        \frac{1}{n}\sum_{j=1}^n \mathds{1}\bigg[0.1 \leq \frac{\sum_{\kappa\in\mathcal{K}}\mathds{1}\big[A_{\kappa}^{(j)}{=}A_{\text{gold}}^{(j)}\big]}{|\mathcal{K}|} \leq 0.9\bigg]
    \end{align}
\end{itemize}
To obtain the above metrics, we consider our task as a classification of each word's head, which is consistent with the computation of our society entropy and is a simplification of the task yet incorporating the diversity notion.

For each diversity metric, we finetune the balancing hyperparameter $\alpha$ in Eqn.~\eqref{eq:modelselectionobjective} with one decimal-place precision based on the performance of the corresponding selected five-individual ensembles.

\begin{figure}[t]
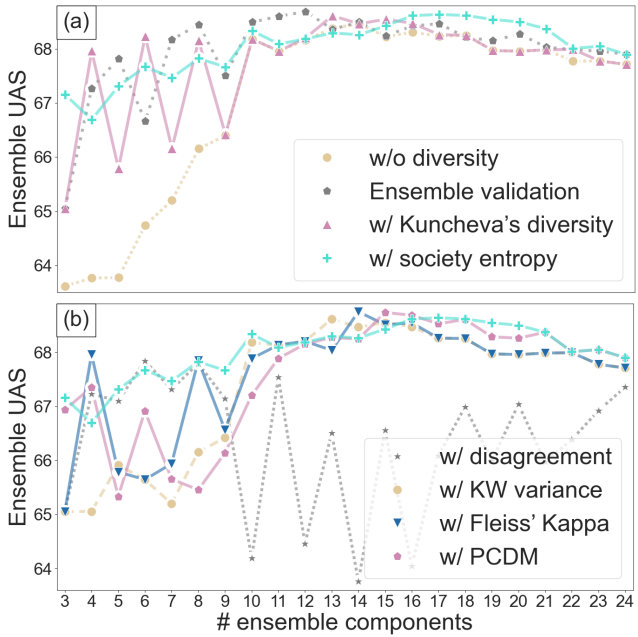

    \centering
    \includegraphics[width=\linewidth]{model_selection_main.pdf}
    \includegraphics[width=\linewidth]{model_selection_appndx.pdf}
    \caption{Ensemble performance by the number of selected ensemble components using different selection strategies. Ensemble selection happens over 25 individuals (five runs for each of CRFAE, NDMV, NE-DMV, L-NDMV, and Sib-NDMV). Results are split into two figures for easier reading.}
    \label{fig:model_selection_main}
\end{figure}

Figure~\ref{fig:model_selection_main} illustrates the performance of ensembles by different numbers of selected individuals. The selection methods include using only individuals' performance (w/o diversity), the ensemble objective given in Eqn.~\eqref{eq:modelselectionobjective} employing different diversity metrics, and the classic forward-stepwise-selection approach (ensemble validation).

\begin{figure*}[t]
    \centering
    \includegraphics[width=\linewidth]{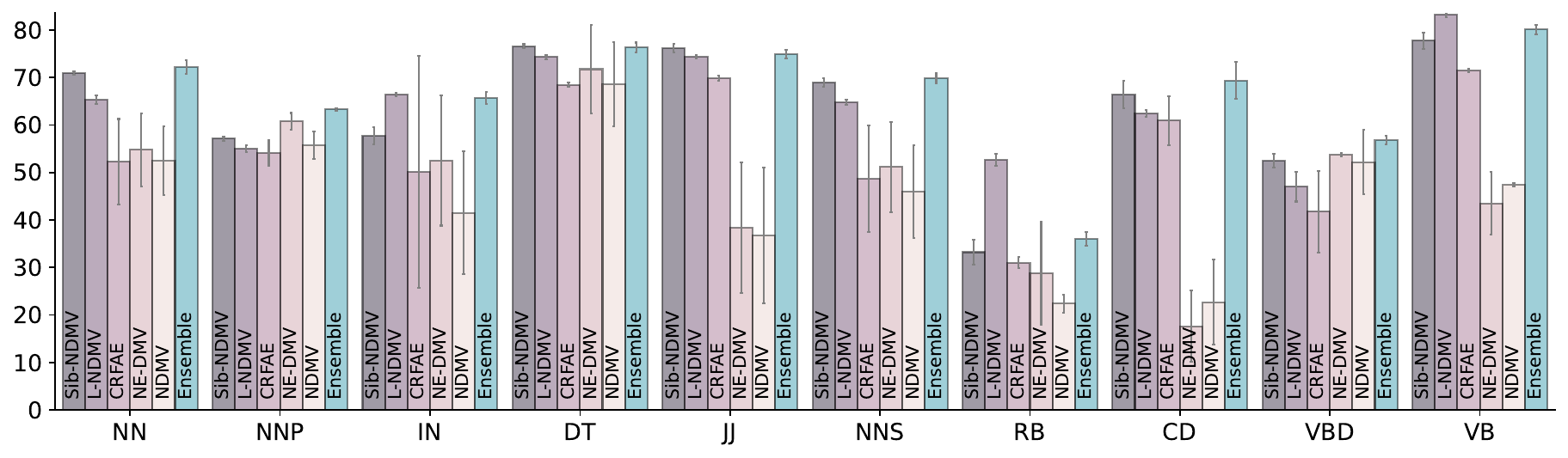}
    \caption{UAS by dependents' POS tags on the WSJ test set. The ten represented POS tags cover around 78\% of the cases.}
    \label{fig:recalls}
\end{figure*}

Our results show that ignoring diversity performs worse than most other methods, underscoring the importance of diversity in ensemble selection. On the other hand, society entropy outperforms the other approaches when a larger number of ensemble components are selected. This is expected, given the higher likelihood of error accumulation with a greater number of weak individuals and that society entropy is the only metric that accounts for error diversity. Moreover, this metric demonstrates significantly more stability than others when we have a small number of individuals. Overall, society entropy performs consistently well, making it the most reliable approach for selecting any number of ensemble components.

Additionally, we present a stability and efficiency analysis in Table~\ref{tab:model_selection_main} for selecting a small number of individuals, as this not only represents a more realistic use case for ensemble selection but also addresses the instability observed in the performance of Kuncheva's diversity and ensemble validation in Figure~\ref{fig:model_selection_main}. To this end, we randomly selected 15 individuals from a group of 25. This random sampling was repeated 30 times to ensure robustness. For each of these 30 sample sets, we applied different selection strategies to choose subsets of 3, 4, and 5 individuals from the 15. Results show that society entropy excels and is more stable than the diversity-free and Kuncheva's diversity approaches, achieving performance comparable to ensemble validation but around~500x faster, making it a practical method.

Finally, we report in Table~\ref{tab:main_results} the performance of selected ensembles. Here, we selected comparable numbers of individuals (five models for Rows~9-12 and six models for Rows~16-17) from all individuals across all runs. Our results are consistent with our analysis, demonstrating that our society entropy is comparable with ensemble validation, outperforming other ensemble-selection methods. Moreover, our society entropy-based selected ensemble outperforms all unselected ensembles and individuals, exhibiting the highest unsupervised dependency parsing performance among all competitors.

\begin{table}[t]
    \centering
    \begin{tabular}{|r l | r |}
        \toprule
        & Method       & Time$_\text{(ms)}$ \\
        \midrule
        \multicolumn{2}{|l|}{Individual inference} & \\
        & CRFAE$^\dag$ & $14.9$             \\
        & NDMV         & $22.3$             \\
        & NE-DMV       & $22.3$             \\
        & L-NDMV       & $56.4$             \\
        & Sib-NDMV     & $114.2$            \\
        & Sib\&L-NDMV  & $174.5$            \\
        \midrule
        \multicolumn{2}{|l|}{Our aggregation} & $19.6$ \\
        \bottomrule
    \end{tabular}
    \caption{Inference time in milliseconds, measured by a single Intel(R) Core(TM) i9-9940X (@3.30GHz) CPU, without GPU, averaged through samples in the WSJ test. $^\dag$Implemented in Java. Others are in Python.}
    \label{tab:run_time}
\end{table}

\subsection{Additional Analysis}

\paragraph{Inference Efficiency.} In Table~\ref{tab:run_time}, we report the inference runtime for every individual, along with the execution time of our aggregation algorithm. While the inference of an ensemble is generally slow as it requires the inference of all the individuals, results show that our proposed aggregation step does not further slow down this process significantly, demonstrating the efficiency of our proposed approach that makes use of \citet{eisner-1996-three}'s algorithm.

\paragraph{Performance by Dependents' Part-of-Speech (POS).}
We would like to investigate the effect of our ensemble approach from a linguistic perspective.
To this end, we report in Figure~\ref{fig:recalls} the breakdown performance of our weighted ensemble and its individuals by the POS tags of the dependents. We note that different individuals have their expertise in different types of dependents. In fact, NE-DMV surpasses the others in NNP and VBD cases, L-NDMV is outstanding in IN, RB, and VB, while Sib-NDMV outperforms other individuals in all other types. In most cases, our ensemble roughly matches the performance of the best individual, outperforming all of them in terms of overall performance.

\paragraph{Appendices.} 
We include more results in the appendix:
(A)ggregation based on F\textsubscript{1} scores,
(B)aseline replication, and
(C)ase studies.

\section{Conclusion}
In this work, we propose a post hoc ensemble approach to unsupervised dependency parsing, equipped with a diversity-aware ensemble-selection method. We emphasize the importance of error diversity alongside expertise diversity and introduce society entropy as a measure that accounts for both. Our experiments demonstrate the effectiveness of our ensemble approach and the critical role of error diversity in ensemble selection.

\paragraph{Future Work.} We identify future directions from algorithmic, linguistic, and machine learning perspectives. On the algorithmic side, it is intriguing to develop aggregation methods for other structures, including non-projective dependency parses or other graph structures in tasks such as drug discovery. For the linguistic aspect, it is promising to investigate cross-task ensembles by aggregating diverse knowledge of different constituency and dependency parsers. From the machine learning perspective, our ensemble selection is limited to classification tasks or those that can be framed as classification. We aim to explore ensemble-selection approaches, incorporating error diversity, for a broader range of problems.

In parallel with our work, \citet{charakorn2024diversity} identify a similar differentiation between expertise diversity (called \textit{specialization}) and general diversity (which includes error diversity) in multi-agent reinforcement learning. It further supports our claim that both aspects of diversity should be taken into account, and suggests a great promise for our approach in multi-agent reinforcement learning.

\section*{Acknowledgments}

We thank all reviewers and chairs for their valuable and constructive comments.
The research is supported in part by the Natural Sciences and Engineering Research Council of Canada (NSERC), the Amii Fellow Program, the Canada CIFAR AI Chair Program, an Alberta Innovates Program, a donation from DeepMind, and the Digital Research Alliance of Canada (alliancecan.ca).
We also thank Hans-Werner Eroms and Timothy J. Osborne for providing advice on dependency parse representations.

\bibliography{aaai25}

\newpage

\appendix

\section{Aggregation Based on F\textsubscript{1} Scores}
\label{sec:f1section}

\paragraph{Formulation.}
In our previous work \citep{shayegh2023ensemble,shayegh2024disco}, we demonstrate the effectiveness of employing $\operatorname{F_1}(T_p, T_r) = \frac{2|C(T_p)\cap C(T_r)|}{|C(T_p)|+|C(T_r)|}$ as the ensemble objective in unsupervised constituency parsing, where $C(T)$ is the set of phrases induced from a parse tree~$T$. We are curious about the performance of $\operatorname{F_1}$ as the ensemble objective in our task of unsupervised dependency parsing.

Adopting such a phrasal F\textsubscript{1} score requires the notion of phrases in the parse structure, which is not directly available in dependency parsing. We view each word and its descendant words as a \textit{dependency phrase}, based on which we can compute the F\textsubscript{1} score.

\begin{figure}[t]
    \centering
    \begin{minipage}[t]{0.05\textwidth}
        \raisebox{.75cm}{(a)}
    \end{minipage}\hfill%
    \begin{dependency}[theme = simple]
        \begin{deptext}[column sep=1.55em]
            \scriptsize Do \& \scriptsize parsers \& \scriptsize dream \& \scriptsize of \& \scriptsize dependency \& \scriptsize trees
            \\
        \end{deptext}
        \deproot{1}{\textsc{root}}
        \depedge{1}{3}{}
        \depedge{3}{2}{}
        \depedge{3}{4}{}
        \depedge{4}{6}{}
        \depedge{6}{5}{}
    \end{dependency}

    \begin{minipage}[t]{0.05\textwidth}
        \raisebox{.75cm}{(b)}
    \end{minipage}\hfill%
    \includegraphics[scale=1]{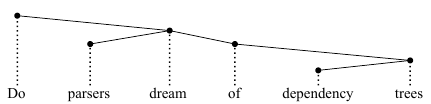}
    \caption{A dependency parse structure represented by directed dependency arcs (a) and DPST (b).}
    \label{fig:dep_phrases}
\end{figure}

We reduce the aggregation in our unsupervised dependency parsing setup to the tree averaging in our previous work~\citep{shayegh2023ensemble,shayegh2024disco}. We may do so by arranging the dependency phrases in a hierarchy~\citep{eroms2000syntax}, which we call a \textit{Dependency-Based Phrase Structure Tree} (DPST; Figure~\ref{fig:dep_phrases}). Notice that DPSTs and traditional arc-based representations are one-to-one corresponding.

\paragraph{Algorithm.} Consider a set of DPSTs given by the ensemble individuals. Our goal is to find an average DPST, which is most similar to the individuals' outputs according to the F\textsubscript{1} score. We develop a dynamic programming (DP) algorithm to achieve this. Notice that a DPST can be non-binary and has a domain-specific constraint that each non-leaf node has to be connected to exactly one leaf node, which we call \textit{one-leaf constraint}.
These make our problem different from that in our previous studies \citep{shayegh2023ensemble,shayegh2024disco}.

Specifically, we would like to find the DPST that maximizes the overall F\textsubscript{1} score against a set of given DPSTs $\{T_1, \cdots, T_k\}$ over an $n$-word sentence, given by
\begin{align}
    T^* &= \underset{T \in \mathcal T}{\operatorname{argmax}} \sum_{k=1}^K \operatorname{F_1}(T, T_k) \\
    &= \underset{T \in \mathcal T}{\operatorname{argmax}} \sum_{k=1}^K \frac{2\big|C(T) \cap C(T_k)\big|}{\big|C(T)\big|+\big|C(T_k)\big|} \label{eq:maxf1withdenom}
\end{align}
where $\mathcal T$ is the set of all possible DPSTs. Given the one-leaf constraint, we have $|C(T)| = n$ for every DPST $T$. This discards the denominator in Eqn.~\eqref{eq:maxf1withdenom} and yields
\begin{align}
    T^* &= \underset{T \in \mathcal T}{\operatorname{argmax}} \sum_{k=1}^K \big|C(T) \cap C(T_k)\big|\\
    &= \underset{T \in \mathcal T}{\operatorname{argmax}} \sum_{c \in C(T)} \underbrace{\sum_{k=1}^K \mathds{1}\big[c \in C(T_k)\big]}_{\displaystyle{{h}(c)}}
\end{align}
showing that F\textsubscript{1}-based aggregation for dependency trees can be reduced to the hit count maximization, similar to constituency trees~\cite{shayegh2023ensemble}. Here, ${h}(c)$ is the number of occurrences of the phrase $c$ in the individuals' outputs.

For our DP algorithm, we define the recursion variable $H_{b:e}$ to be the total hit count of the best substructure over the subsequence $b{:}e$ of the sentence (starting from the $b$th word and ending before the $e$th word)
\begin{align}
    H_{b:e} = \max_{T \in \mathcal{T}_{b:e}} \sum_{c \in C(T)} {h}(c)
\end{align}
where $\mathcal{T}_{b:e}$ is the set of all possible substructures over $b{:}e$. Specially, we define $H_{b:b}=0$ for algorithmic convenience. 

For recursion, we notice that the branching of a node may not be binary, and that a non-binary branching can be converted to binary branches by inserting placeholder nodes.
Therefore, we define two additional DP variables:
$H_{b:e}^{\text{excl}}$ referring to the total hit count when the substructure over span $b:e$ is a placeholder, and $H_{b:e}^{\text{incl}}$ when the substructure over $b:e$ is an actual node in the tree (i.e., forming a dependency phrase).

The recursion formulas are
\begin{align}
    H_{b:e}^{\text{excl}} &= \max_{b < j < e}
    (H_{b:j} + H_{j:e})
    \label{eq:Hexcl}\\
    H_{b:e}^{\text{incl}} &= \max_{b \leq j < e}
    (H_{b:j} + {h}(b{:}e) + H_{j+1:e})
    \label{eq:Hincl}
\end{align}
Note that \eqref{eq:Hincl} suggests the substructure over span $b:e$ is an actual node in the tree. Based on the one-leaf constraint, the $j$th word does not appear in its left branch $b:j$ or right branch $j+1:e$.

The best substructure over $b{:}e$ either includes or excludes the phrase $b{:}e$ itself.
\begin{align}
    H_{b:e} = \max\big\{H_{b:e}^{\text{incl}}, H_{b:e}^{\text{excl}}\big\}
    \label{eq:Hoverall}
\end{align}

The recursion terminates once we have computed $H_{1:n+1}$ for a length-$n$ sentence. To construct $T^*$, we backtrack the best corresponding substructures along our recursion.

We analyze the time complexity of our algorithm. For a length-$n$ sentence, the number of DP variables is $\mathcal{O}(n^2)$, because we need to compute $H$, $H^{\text{incl}}$, and $H^{\text{excl}}$ for $\binom{n}{2}$-many possible subsequences in the sentence. To compute the value of each variable, we need to go through Eqn.~\eqref{eq:Hexcl} and Eqn.~\eqref{eq:Hincl}, each having a complexity of $\mathcal{O}(n)$ as we need to find the best~$j$. Eqn.~\eqref{eq:Hoverall} does not add to the complexity, and the overall time complexity of our algorithm is $\mathcal{O}(n^3)$.

\begin{table}[t]
    \centering
    \begin{tabular}{|r l | l | l |}
        \toprule
        & Method                               & UAS                    & F\textsubscript{1} \\
        \midrule
        \multicolumn{2}{|l|}{Ensemble individuals} & & \\
        1 & CRFAE                                & $53.0_{\pm4.5}$        & $50.4_{\pm4.2}$    \\
        2 & NDMV                                 & $48.1_{\pm3.6}$        & $52.0_{\pm1.8}$    \\
        3 & NE-DMV                               & $51.0_{\pm3.4}$        & $53.6_{\pm0.9}$    \\
        4 & L-NDMV                               & $62.4_{\pm0.6}$        & $63.4_{\pm0.5}$    \\
        5 & Sib-NDMV                             & $64.3_{\pm0.2}$        & $64.9_{\pm0.3}$    \\
        \midrule
        \multicolumn{2}{|l|}{Our ensembles} & & \\
        6 & F\textsubscript{1}-based aggregation & $59.2_{\pm2.1}$        & $64.5_{\pm0.9}$    \\
        7 & UAS-based aggregation                & ${{65.7}}_{\pm0.9}$ & $63.1_{\pm0.5}$    \\
        \midrule
        \multicolumn{2}{|l|}{Our weighted ensembles} & & \\
        8 & F\textsubscript{1}-based aggregation & $62.0_{\pm1.6}$        & $\bf{66.6}_{\pm0.8}$    \\
        9 & UAS-based aggregation                & ${\bf{66.6}}_{\pm0.4}$ & $63.7_{\pm0.4}$    \\
        \bottomrule
    \end{tabular}
    \caption{Mean and standard deviation of performance across five runs, evaluated on the WSJ test set.}
    \label{tab:f1results}
\end{table}

\paragraph{Results.}
Table~\ref{tab:f1results} shows the performance of our F\textsubscript{1}-based ensemble approach for unsupervised dependency parsing. We include both UAS (being the standard metric) and the phrasal F\textsubscript{1} score (aligning with the ensemble criterion) for evaluation. For the F\textsubscript{1} metric, we use the corpus-level variant.

We see that F\textsubscript{1}-based aggregation is less successful than our main approach, UAS-based aggregation (\S\ref{sec:aggs}): both vanilla and weighted F\textsubscript{1}-based ensembles (Rows~6 and~8) fail to outperform the best individual under the standard UAS metric, although the weighted ensemble (Row~8) yields an improvement of 1.7 points in F\textsubscript{1}. By contrast, UAS-based ensembles  (Rows~7 and~9) consistently outperform all individuals under the standard UAS metric.

Interestingly, Rows 6--9 show that F\textsubscript{1} and UAS are not perfectly correlated, as optimizing one does not necessarily improve the other.

\section{Baseline Replication}
\label{sec:replication}
We report in Table~\ref{tab:QuotedResults} the quoted and replicated results for our ensemble components. Since CRFAE, NDMV, and NE-DMV exhibit unstable and subpar performance compared with their quoted results, we train 20 runs each and select the five top-performing runs based on validation performance. As demonstrated, our results are close to previously reported ones, providing a fair foundation for the subsequent experiments and analyses.

\begin{table}[t]
    \centering
    \begin{tabular}{| l | l | l | l |}
        \toprule
        \multirow{2}{*}{Method} & \multirow{2}{*}{Quoted} & \multicolumn{2}{c|}{Replication} \\
        &        & 20 runs         & Top-5 runs      \\
        \midrule
        CRFAE\tablefootnote{Quoted from \citet{cai2017crf} and replicated using \url{https://github.com/shtechair/CRFAE-Dep-Parser}\label{footnote:crfaefootnote}} & $55.7$ & $45.0_{\pm5.2}$ & $53.0_{\pm4.5}$\\
        NDMV\tablefootnote{Quoted from \citet{jiang2016unsupervised} and replicated using \url{https://github.com/LouChao98/neural_based_dmv}\label{footnote:neural_based_dmv}} & $47.0$ & $43.4_{\pm4.9}$ & $48.1_{\pm3.6}$\\
        NE-DMV\footnoteagain{footnote:neural_based_dmv} & $52.5$ & $43.8_{\pm5.2}$ & $51.0_{\pm3.4}$ \\
        L-NDMV\tablefootnote{Quoted from \citet{han-etal-2017-dependency} and replicated using \url{https://github.com/sustcsonglin/second-order-neural-dmv}\label{footnote:lndmvfootnote}} & $59.5$ & -- & $62.4_{\pm0.6}$\\
        Sib-NDMV\tablefootnote{Quoted from \citet{yang-etal-2020-second} and replicated using \url{https://github.com/sustcsonglin/second-order-neural-dmv}\label{footnote:second_order_repo}} & $64.5$ & -- & $64.3_{\pm0.2}$ \\
        Sib\&L-NDMV\footnoteagain{footnote:second_order_repo} & $67.5$ & -- & $67.9_{\pm0.1}$ \\
        \bottomrule
    \end{tabular}
    \caption{The UAS metric for baseline models on the WSJ test set. We report the average and standard deviation across various runs. The quoted results are from the original papers.
    }
    \label{tab:QuotedResults}
\end{table}

\section{Case Studies}
\label{sec:casestudies}

Table~\ref{tab:casestudies} presents three cases illustrating how our ensemble selection performs. We compare our society entropy-based ensemble selection with two baselines: ensemble selection without considering diversity and ensemble selection using Kuncheva's diversity~\citep{kuncheva2003measures}. Recall that we have five runs for five models (CRFAE,\footnoteagain{footnote:crfaefootnote} NDMV,\footnoteagain{footnote:neural_based_dmv} NE-DMV,\footnoteagain{footnote:neural_based_dmv} L-NDMV,\footnoteagain{footnote:lndmvfootnote} and Sib-NDMV\footnoteagain{footnote:second_order_repo}), totaling 25 candidate individuals. In this case study, we set the number of selected individuals to be three for presentation purposes.

Eventually, all strategies select one run of L-NDMV (denoted by L-NDMV\textsubscript{1}) and one run of Sib-NDMV, while the third selection differs. Both baselines choose another L-NDMV run (denoted by L-NDMV\textsubscript{2}) with a UAS performance of around~63\%, while our method picks NE-DMV with a much lower performance of around~52\%. Intriguingly, in cases where both L-NDMV\textsubscript{1} and NE-DMV err, NE-DMV produces errors that are diverse from those of L-NDMV\textsubscript{1}, resulting in scattered errors among the selected components and making it easier for the ensemble to eliminate the errors. By contrast, two L-NDMV runs tend to agree with each other, which is expected. This misdirects the ensemble when they make identical errors.

Take Table~\ref{tab:casestudy1} as an example. L-NDMV\textsubscript{1} assigns the wrong labels at positions 1, 2, 4, and 5, while Sib-NDMV assigns the correct labels. 
At the above positions, L-NDMV\textsubscript{2} makes the exact same errors, cementing them in the ensemble output. 
By contrast, NE-DMV produces different errors at positions 1 and 4 (known as error diversity), scattering the errors and letting the ensemble to predict the true labels. 
Further, NE-DMV assigns the correct labels at positions 2 and 5, while having mistakes in other positions like 7, providing its unique expertise (known as expertise diversity) different from that of L-NDMV\textsubscript{1}.

These examples show the importance of both error diversity and expertise diversity in eliminating errors and achieving a better ensemble performance.

\addtocounter{footnote}{1}\footnotetext{Our two baselines include ensemble selection without considering diversity and ensemble selection using Kuncheva's diversity~\citep{kuncheva2003measures}.\label{footnote:ensembleselectionbaselines}}
\addtocounter{footnote}{1}\footnotetext{Our ensemble selection uses society entropy as the diversity metric.\label{footnote:ensembleselectionwithsocietyentropy}}

\setlength{\tabcolsep}{1mm}
\renewcommand{\arraystretch}{1.5}

\begin{table*}[t]
    \centering
    \begin{subtable}{\textwidth}
        \begin{tabular}{| l | l | c c c c c c c c c c c |}
            \toprule
            \multicolumn{2}{| c |}{Positions} & 1 & 2 & 3 & 4 & 5 & 6 & 7 & 8 & 9 & 10 & 11 \\
            \multicolumn{2}{| c |}{Sentence} & In & a & Centennial & Journal & article & Oct. & 5 & the & fares & were & reversed \\
            % \hline
            \multicolumn{2}{| c |}{True heads} & were & article & article & article & In & were & Oct. & fares & were & \textsc{root} & were \\
            \midrule
            \multirow{2}{*}{\makecell[l]{Common \\ individuals}}              & Sib-NDMV & were & article & Journal & article & In & were & Oct. & fares & Oct. & \textsc{root} & were \\ \cline{2-13}
            & L-NDMV\textsubscript{1} & the & Centennial & Journal & In & Journal & article & Oct. & fares & were & \textsc{root} & were \\
            \bottomrule
            \toprule
            Baselines\footnoteagain{footnote:ensembleselectionbaselines}      & L-NDMV\textsubscript{2}          & the      & Centennial         & Journal         & In         & Oct.    & Journal      & Oct.      & fares       & were      & \textsc{root}               & were      \\
            \hline
            \multicolumn{2}{| c |}{Ensemble} & the & Centennial & Journal & In & Journal & article & Oct. & fares & were & \textsc{root} & were \\
            \bottomrule
            \toprule
            Ours\footnoteagain{footnote:ensembleselectionwithsocietyentropy} & NE-DMV & Oct. & article & Journal & a & In & 5 & were & fares & 5 & \textsc{root} & were \\
            \hline
            \multicolumn{2}{| c |}{Ensemble} & were & article & Journal & article & In & were & Oct. & fares & were & \textsc{root} & were \\
            \bottomrule
        \end{tabular}
        \subcaption{}
        \label{tab:casestudy1}
    \end{subtable}
    \vspace{.3cm}\\
    \begin{subtable}{\textwidth}
        \begin{tabular}{| l | l | c c c c c c c c |}
            \toprule
            \multicolumn{2}{| c |}{Positions} & 1 & 2 & 3 & 4 & 5 & 6 & 7 & 8 \\
            \multicolumn{2}{| c |}{Sentence} & Does & this & signal & another & Black & Monday & is & coming \\
            \multicolumn{2}{| c |}{True heads} & \textsc{root} & Does & Does & Monday & Monday & is & signal & is \\
            \midrule
            \multirow{2}{*}{\makecell[l]{Common \\ individuals}} & Sib-NDMV & \textsc{root} & Does & Does & Black & Monday & is & signal & is \\ \cline{2-10}
            & L-NDMV\textsubscript{1} & is & Does & Does & signal & signal & Black & \textsc{root} & is \\
            \bottomrule
            \toprule
            Baselines\footnoteagain{footnote:ensembleselectionbaselines} & L-NDMV\textsubscript{2} & is & Does & Does & signal & Monday & signal & \textsc{root} & is \\
            \hline
            \multicolumn{2}{| c |}{Ensemble} & is & Does & Does & signal & Monday & is & \textsc{root} & is \\
            \bottomrule
            \toprule
            Ours\footnoteagain{footnote:ensembleselectionwithsocietyentropy} & NE-DMV & \textsc{root} & signal & Does & Monday & Monday & is & signal & is \\
            \hline
            \multicolumn{2}{| c |}{Ensemble} & \textsc{root} & Does & Does & Monday & Monday & is & signal & is \\
            \bottomrule
        \end{tabular}
        \subcaption{}
        \label{tab:casestudy2}
    \end{subtable}
    \vspace{.3cm}\\
    \begin{subtable}{\textwidth}
        \begin{tabular}{| l | l | c c c c c |}
            \toprule
            \multicolumn{2}{| c |}{Positions} & 1 & 2 & 3 & 4 & 5 \\
            \multicolumn{2}{| c |}{Sentence} & More & Elderly & Maintain & Their & Independence \\
            \multicolumn{2}{| c |}{True heads} & Maintain & More & \textsc{root} & Independence & Maintain \\
            \midrule
            \multirow{2}{*}{\makecell[l]{Common \\ individuals}} & Sib-NDMV & Maintain & More & \textsc{root} & Independence & Maintain \\\cline{2-7}
            & L-NDMV\textsubscript{1} & Elderly & Maintain & \textsc{root} & Independence & Maintain \\
            \bottomrule
            \toprule
            Baselines\footnoteagain{footnote:ensembleselectionbaselines} & L-NDMV\textsubscript{2} & Elderly & Maintain & \textsc{root} & Independence & Maintain \\
            \hline
            \multicolumn{2}{| c |}{Ensemble} & Elderly & Maintain & \textsc{root} & Independence & Maintain \\
            \bottomrule
            \toprule
            Ours\footnoteagain{footnote:ensembleselectionwithsocietyentropy} & NE-DMV & Maintain & More & \textsc{root} & Independence & Maintain \\
            \hline
            \multicolumn{2}{| c |}{Ensemble} & Maintain & More & \textsc{root} & Independence & Maintain \\
            \bottomrule
        \end{tabular}
        \subcaption{}
        \label{tab:casestudy3}
    \end{subtable}
    \caption{Case studies.}
    \label{tab:casestudies}
\end{table*}

\end{document}